\documentclass{svproc}
\usepackage{url}

\usepackage{amsmath}
\usepackage{subfigure}
\usepackage{graphicx}
\usepackage{hyperref}

\begin{document}
\mainmatter              
\title{Forest Proximities for Time Series}
\titlerunning{Forest Proximities for Time Series}  
%

\author{Ben Shaw\inst{1} \and Jake Rhodes\inst{2} \and
Soukaina Filali Boubrahimi\inst{3} \and Kevin R. Moon\inst{4}}
\authorrunning{Shaw et al.} 
%
\tocauthor{First Author, Second Author, Third Author and Fourth Author}
\institute{Utah State University, Logan UT, USA,\\
\email{ben.shaw@usu.edu}
\and
Brigham Young University, Provo UT, USA,\\
\email{rhodes@stat.byu.edu}
\and
Utah State University, Logan UT, USA,\\
\email{soukaina.boubrahimi@usu.edu}
\and
Utah State University, Logan UT, USA,\\
\email{kevin.moon@usu.edu}
}


\maketitle              

\begin{abstract}
RF-GAP has recently been introduced as an improved random forest proximity measure. In this paper, we present PF-GAP, an extension of RF-GAP proximities to proximity forests, an accurate and efficient time series classification model. We use the forest proximities in connection with Multi-Dimensional Scaling to obtain vector embeddings of univariate time series, comparing the embeddings to those obtained using various time series distance measures. We also use the forest proximities alongside Local Outlier Factors to investigate the connection between misclassified points and outliers, comparing with nearest neighbor classifiers which use time series distance measures. We show that the forest proximities seem to exhibit a stronger connection between misclassified points and outliers than nearest neighbor classifiers.
\keywords{Proximity Forests, time series classification, random forest proximities, time series outlier detection, time series embedding methods.}
\end{abstract}

\section{Introduction}

Random forests (RF)~\cite{RF} is an important machine learning algorithm that is still widely used today for both classification and regression problems (see the citations in~\cite{RFGAP} for some recent examples). Supervised pairwise similarity measures between data points that are constructed from the RF, known as RF proximities, have been defined in several ways~\cite{RF,RFoob,RFpbk,RFProxIH}.
RF proximities have been used in many applications, including outlier detection \cite{RF}, visualization \cite{RFoob}, multiview learning \cite{RFProxIH}, and defining a kernel matrix for support vector machines \cite{RFpbk}. A recently defined RF proximity measure, RF-Geometry and Accuracy Preserving (GAP) proximities~\cite{RFGAP} captures the data geometry learned by the RF in the sense that a proximity-weighted neighbor classifier or regressor both theoretically and empirically output the same out-of-bag predictions as the original RF. This property translated to improvements in multiple applications including dimensionality reduction for visualization, missing data imputation, and outlier detection~\cite{RFGAP,RFPHATE}. 

Despite the utility of random forests, including random forest proximities, random forests do not always perform well on time series data \cite{PF}. Proximity forests~\cite{PF} were recently introduced to fill this gap by extending an RF-like algorithm to time-series data. However, proximities for this new forest algorithm have not been previously explored. We introduce a definition of proximities for proximity forests to extend the applicability of proximity forests for time series data in much the same way as forest proximities extend the applicability of random forests for other data types. Due to their geometry-preserving capabilities, we specifically extend the RF-GAP proximities to proximity forests, and call the resulting proximities PF-GAP. 

While there are many potential applications of PF-GAP, we focus mainly on time series outlier detection and the connection between outliers and misclassified points. We show that PF-GAP outperforms other approaches that use different, common time series distance measures. We also demonstrate how PF-GAP can be used to obtain vector embeddings in a supervised manner. Our contributions can be summarized as follows: (1) we \href{https://sites.google.com/view/forest-proximities}{implement} GAP proximities for the proximity forest model;\footnote{See https://sites.google.com/view/forest-proximities for code used to produce the experiments. See also https://github.com/KevinMoonLab/PF-GAP/tree/main for PF-GAP source code.} (2) we demonstrate the utility of time series proximities in supervised visualization and within-class outlier detection; (3) we conduct an experiment using 64 datasets from the UCR 2018 archive \cite{UCRArchive2018} which suggests a stronger connection between outlier time series and misclassified points than with nearest neighbor classifiers. 



This paper is organized as follows. Section \ref{sec:Background} gives background information on the proximity forest model and random forest proximities. Additionally, the latter part of section \ref{sec:Background} describes the meaning of time series outlier detection as it it used in this paper. Section \ref{sec:methods} describes the application of GAP proximities to the proximity forest model, distances induced by the proximities, as well as the method by which within-class outlier scores are obtained. Section \ref{sec:Experiments} provides both the main experimental result of the paper as well as a visual result depicting the use of GAP proximities for time series visualization. Section \ref{sec:Discussion} provides a brief discussion of limitations, and we provide concluding remarks in section \ref{sec:Conclusion}.

\section{Background}
\label{sec:Background}
\subsection{Proximity Forests}
\label{sec:PF}

Proximity Forests were proposed in 2018 as a comparatively efficient and accurate time series classification model \cite{PF}. Early experiments by its creators have shown that proximity forests generally outperform time series classifiers such as BOSS-VS \cite{BOSSVS}, WEASEL \cite{WEASEL}, and ElasticEnsemble \cite{ElasticEnsemble} in terms of accuracy, and that the accuracy is generally comparable to classifiers such as COTE \cite{COTE}.  

A proximity forest is an ensemble of proximity trees, with a final prediction being made by voting across trees, as in RF. 
Where typical decision trees split at a particular node based on features, a proximity tree at a particular node creates a number of branches equal to the number of classes at the node, assigning a random data point from each class to a branch. Such data points are known as ``exemplars.'' Subsequently, data at the node in question traverse the branch according to which exemplar it is nearest to based on a pre-defined distance measure. 

Initially, the exemplars are randomly sampled, and the collection of sampled exemplars is called a candidate split. A total of $r$ candidate splits are randomly generated, where $r$ is a user-specified hyperparameter. The quality of each candidate split is assessed by measuring the Gini purity of the potential child nodes. As the $r$ parameter is increased, the model has a better chance of obtaining the ideal splits at each node. However, increasing the $r$ parameter also increases the computational resources required to train the proximity tree.

A node is called a leaf node if all data labels within the node are the same, meaning the node is pure. If the node is not pure, branches are created for each class of data at the current node and the tree is grown. The recursive algorithm terminates either when there are no subtrees to create due to the obtained purity or when a specified tree depth is reached.


The proximity trees are intended to be independent of one another, due to the randomness induced by the chosen exemplars and the random choice of distance measure. If no user-specified distance measure is given, each tree uses a random selection from a list of possible distance measures. For the original Java implementation which we use, these nine distance measures are Dynamic Time Warping (DTW), Derivative DTW (DDTW), Weighted DTW (WDTW), Weighted DDTW (WDDTW), Time Warp Edit distance (TWE), Euclidean Distance (ED), Longest Common Subsequence (LCSS), Move-Split-Merge (MSM), and Edit distance with Real Penalty (ERP). See \cite{bakeoff,ElasticEnsemble,Comparison} for descriptions and comparisons.


\subsection{Random Forest Proximities}

Recall that a given tree in a random forest is typically trained on a bootstrap sample. Points that are in the bootstrap sample are called ``in-bag" while the excluded points are considered out-of-bag (OOB). Given a trained random forest, the original proximity between two observations $i$ and $j$ was defined as the proportion of trees in the random forest that the two points end up in the same terminal (leaf) node \cite{RF}:
\begin{equation}{\label{Por}}
    p_{Or}(i,j) = \frac{1}{T} \sum_{t=1}^T I(j \in v_i(t)),
\end{equation}
where $T$ is the number of trees in the forest, $v_i(t)$ is the set of indices of observations that end up in the same terminal node as $x_i$ in tree $t$, and $I$ is the indicator function. However, it was shown in~\cite{RFGAP} that this definition distorts the learned RF geometry as it does not take into account the OOB or in-bag status of the points nor the number of in-bag points in the shared terminal node. This weakness is overcome by RF-GAP. The RF-GAP proximity measure, denoted as $p_{GAP}$, relates the similarity between an observation $i$ and an observation $j$ while accounting for the number of training points (in-bag points) in the shared leaf node.  

The formula, introduced in \cite{RFGAP}, is defined as follows. Let $B(t)$ be the multiset of (potentially repeated) indices of in-bag observations of tree $t$. Define $J_i(t) \bigcap v_i(t)$: in-bag observations which share the same leaf/terminal node as index $i$ of tree $t$. $M_i(t)$ is defined to be the multiset $J_i(t)$ but including in-bag repetitions. Let $c_j(t)$ be the multiplicity of index $j$ in the sample. Finally, let $S_i$ be the set of trees for which the index $i$ is out-of-bag. Then for observations $i$ and $j$, their proximity measure is defines as
\begin{equation}{\label{pGAP}}
    p_{GAP}(i,j) = \frac{1}{|S_i|} \sum_{t \in S_i} \frac{c_j(t) \cdot I(j \in J_i(t))}{|M_i(t)|}.
\end{equation}

The proximities introduced by RF-GAP have advantages compared to other random forest proximities~\cite{RFGAP}. For example, $p_{Or}$ proximities provide a biased estimate of the random forest prediction function when used in a proximity-weighted neighbor classifier or regressor, overemphasizing class separation compared to actual random forest predictions. Another RF proximity definition was defined that only compares OOB points within a given tree \cite{RFoob}. However, this measure also fails to reconstruct the RF predictions \cite{RFGAP}. The advantages of $p_{GAP}$ over other forest proximities thus lead us to adapt them for proximity forests.

\subsection{Time Series Outlier Detection}

Outlier detection for time series can mean different things, and several methods of detecting outliers for time series exist \cite{outliersurvey,outliersurvey2,filali2022mining}. In this paper, we consider time series outlier detection to be the identification of an anomalous time series (rather than a specific interval within a series), given a set of multiple time series. Most work on anomaly detection in time series attempts to identify anomalous points within a time series at specific times \cite{outliersurvey2}.

Random forest proximities allow us to calculate within-class outlier scores for each data point in a given dataset \cite{RFGAP}. PF-GAP for outlier detection can be applied by examining which time series have comparatively high outlier scores using the proximities directly. In addition, a method known as Local Outlier Factors (LOF) can categorize points as inliers and outliers based on pairwise distances \cite{LOF}. LOF computes the average distance of a given point (a time series, in our case) to its $k$ nearest neighbors, where $k$ is a hyperparameter. Subsequently, points are categorized as inliers or outliers based on the local density around a given point. We apply LOF using forest proximities as well as pairwise distances obtained using other time series distance measures.

\section{Methods}
\label{sec:methods}

\subsection{PF-GAP and Pairwise Dissimilarity}

As originally implemented, proximity forests do not use bootstrap sampling; each proximity tree is trained on the full set of training data. Thus, all samples are considered in-bag, making bootstrap sampling essential for the non-trivial introduction of $p_{GAP}$ for proximity forests. Therefore, we modify the implementation of proximity forests to accommodate bootstrap sampling. The PF-GAP proximities are then computed in the same manner as the RF-GAP proximities using the proximity forest ensemble of trees and the in-bag and out-of-bag samples.

The computational complexity of this approach is derived from the computational complexities of both the proximity forest classifier and of $p_{GAP}$. The computational complexity of the proximity forest classifier is claimed to be $\mathcal{O}(\log(n) \cdot l^2)$, where $n$ is the number of time series, and $l$ is the length of each time series \cite{PF}. The computational complexity of computing $p_{GAP}(i,j)$ for all indices $1 \leq i,j \leq n$ is $\mathcal{O}(n^2)$. Thus, the computational complexity of PF-GAP scales quadratically with the number of time series and the length of the time series in an additive manner.

Since the computation of $p_{GAP}$ depends only on the forest/tree structure, the same theoretical properties of RF-GAP carry over to PF-GAP. In particular, the proximity-weighted classification property holds, that is, the proximity forest out-of-bag classification prediction can be reconstructed by a weighted-majority vote using $p_{GAP}$ proximities as weights \cite{RFGAP}. Another property we note is $\sum_{j} p_{GAP}(i,j) = 1$  \cite{RFGAP}.

The $p_{GAP}$ proximities can be used to construct pairwise dissimilarity. However, the $p_{GAP}$ proximities are generally not symmetric. Thus, for use in constructing pairwise dissimilarity, we symmetrize the $p_{GAP}$ proximities: 
\begin{equation}{\label{prox}}
    P(i,j) = \frac{1}{2} \left( p_{GAP}(i,j) + p_{GAP}(j,i) \right).
\end{equation}
Since $0 \leq p_{GAP}(i,j) \leq 1$, we obtain pairwise \textit{dissimilarity} $d_{ij}$: 
\begin{equation}{\label{dissim}}
    d_{ij} = 1 - P_{ij}.
\end{equation}
Since $\sum_{j} p_{GAP}(i,j) = 1$, the RF-GAP proximities are generally small. Therefore, to obtain larger differences in distances, we define our pairwise distances as
\begin{equation}{\label{dissim2}}
    d_{ij} = (1 - P_{ij})^2.
\end{equation}

We call the distance measure defined by equation \ref{dissim2} the DGAP distance.

\subsection{Outlier Detection}

We apply the PF-GAP proximities to perform outlier detection. Outlier time series data points are loosely defined as time series that are comparatively dissimilar to the main body of the data. We can also describe outlier points relative to the class in which the points belong. Forest proximities can be used to give intra-class outlier scores as follows. For observation $i$, a raw outlier measure score is defined as \cite{RF}:
\begin{equation}{\label{pouts}}
    \sum_{j \in \text{class(i)}} \frac{N}{P(i,j)^2},
\end{equation}
where $N$ is the size of the dataset, and where $P(i,j)$ is given in Eq. \ref{prox}. After the raw outlier scores are computed, we can compute the medians and mean absolute deviations of the raw scores within each class. Outlier scores are then obtained by subtracting the class-dependent median from the raw scores, and dividing by the class-dependent mean absolute deviation.


In our experiments on outlier detection, we use LOF for easier comparisons with commonly used dissimilarities for time series. For a given dataset, we compute pairwise distances for each of the nine time series distance measures mentioned previously. For each distance measure, we create a $1$-nearest neighbor classifier, using the classifier to obtain predictions for each time series. Next, we use the pairwise distances with LOF (using $5$ nearest neighbors) to predict outliers. Ideally, a time series is considered an outlier if and only if it is misclassified.

\section{Experiments}
\label{sec:Experiments}

We wish to quantify the relationship between outlier time series and misclassified time series. For each selected training dataset in the UCR 2018 archive \cite{UCRArchive2018}, we train 10 distinct classifiers corresponding to both the DGAP distance as well as the nine time series distances mentioned previously. In tandem, each distance measure is used to compute pairwise distances which are subsequently used with LOF (using five nearest neighbors) to predict outliers. For each of the 10 distances, F1 scores are obtained: we treat a correctly classified point as a true inlier, using the output of the associated LOF model as an inlier/outlier prediction. Thus, a true positive is a correctly classified point which is an inlier, a true negative is a misclassified point labeled an outlier, a false negative is a correctly classified point which is labeled an outlier, and a false positive is a misclassified point which is labeled an inlier. For the DGAP distance, the associated classifier is a PF model with $r=5$ and 11 trees. For the remaining distances, the associated classifiers are 1-NN classifiers which use the respective time series distances. The F1 scores for each dataset and distance measure are shown in Table \ref{tab:64results}. 

A total of 64 datasets from the UCR archive were used, though there are 128 datasets in the archive. Some were removed due to missing data. However, the majority of removals occurred due to the length of time required to train $1$-NN distance classifiers for 9 distinct distance measures: for some datasets, this task appeared to require multiple days of computing using CPU.\footnote{This highlights another advantage of the more efficient proximity forest model.} Our selection consisted of datasets for which the $1$-NN classifiers could be computed within a total of 18 minutes.


\begin{table}

\caption{F1 scores for each dataset and distance measure}
\begin{center}
\resizebox{\textwidth}{!}{
\begin{tabular}{lcccccccccc}
\hline
UCR Dataset  & DTW \hspace{1mm} & DDTW \hspace{1mm} & TWE \hspace{1mm} & WDTW \hspace{1mm} & WDDTW \hspace{1mm} & Euclidean \hspace{1mm} & LCSS \hspace{1mm} & MSM \hspace{1mm} & ERP \hspace{1mm} & DGAP\\
\hline
BirdChicken & 0.76 & 0.8 & 0.89 & 0.71 & 0.82 & 0.82 & 0.75 & 0.82 & 0.82 & 1.0\\
\hline
Lightning7 & 0.77 & 0.67 & 0.84 & 0.85 & 0.72 & 0.78 & 0.78 & 0.84 & 0.83 & 1.0\\
\hline
InsectWingbeatSound & 0.47 & 0.31 & 0.65 & 0.58 & 0.56 & 0.65 & 0.67 & 0.66 & 0.58 & 0.99\\
\hline
Adiac & 0.74 & 0.71 & 0.76 & 0.74 & 0.74 & 0.74 & 0.08 & 0.75 & 0.75 & 1.0\\
\hline
FreezerSmallTrain & 0.88 & 0.9 & 0.85 & 0.88 & 0.94 & 0.96 & 0.86 & 0.89 & 0.89 & 1.0\\
\hline
MedicalImages & 0.76 & 0.68 & 0.8 & 0.77 & 0.68 & 0.76 & 0.71 & 0.77 & 0.79 & 1.0\\
\hline
Rock & 0.75 & 0.69 & 0.79 & 0.75 & 0.92 & 0.75 & 0.79 & 0.79 & 0.79 & 1.0\\
\hline
CinCECGTorso & 0.64 & 0.73 & 0.93 & 0.93 & 0.99 & 0.92 & 0.96 & 0.92 & 0.8 & 1.0\\
\hline
WordSynonyms & 0.75 & 0.77 & 0.88 & 0.74 & 0.81 & 0.75 & 0.79 & 0.86 & 0.84 & 1.0\\
\hline
GunPoint & 0.76 & 0.85 & 0.95 & 0.8 & 0.89 & 0.89 & 0.68 & 0.89 & 0.89 & 1.0\\
\hline
FreezerRegularTrain & 0.86 & 0.86 & 0.92 & 0.87 & 0.85 & 0.85 & 0.75 & 0.91 & 0.91 & 1.0\\
\hline
FaceFour & 0.79 & 0.85 & 0.93 & 0.82 & 0.75 & 0.77 & 0.93 & 0.91 & 0.88 & 1.0\\
\hline
Car & 0.72 & 0.8 & 0.9 & 0.7 & 0.86 & 0.81 & 0.49 & 0.89 & 0.85 & 0.99\\
\hline
GunPointMaleVersusFemale & 0.93 & 0.86 & 1.0 & 0.92 & 0.87 & 0.97 & 1.0 & 1.0 & 0.96 & 1.0\\
\hline
OliveOil & 0.9 & 0.78 & 0.91 & 0.92 & 0.78 & 0.91 & 0.29 & 0.91 & 0.91 & 0.98\\
\hline
MiddlePhalanxTW & 0.71 & 0.75 & 0.71 & 0.72 & 0.76 & 0.72 & 0.58 & 0.7 & 0.69 & 0.98\\
\hline
SonyAIBORobotSurface2 & 0.88 & 0.84 & 0.92 & 0.86 & 0.87 & 0.85 & 0.82 & 0.9 & 0.92 & 1.0\\
\hline
FacesUCR & 0.89 & 0.87 & 0.98 & 0.89 & 0.84 & 0.86 & 0.85 & 0.97 & 0.96 & 1.0\\
\hline
PowerCons & 0.81 & 0.75 & 0.95 & 0.83 & 0.77 & 0.96 & 0.94 & 0.95 & 0.93 & 1.0\\
\hline
ProximalPhalanxTW & 0.85 & 0.85 & 0.85 & 0.85 & 0.85 & 0.85 & 0.08 & 0.85 & 0.85 & 0.99\\
\hline
SyntheticControl & 0.94 & 0.76 & 0.99 & 0.94 & 0.74 & 0.96 & 0.86 & 0.98 & 1.0 & 1.0\\
\hline
MiddlePhalanxOutlineCorrect & 0.83 & 0.84 & 0.87 & 0.83 & 0.84 & 0.87 & 0.79 & 0.87 & 0.87 & 0.99\\
\hline
MiddlePhalanxOutlineAgeGroup & 0.85 & 0.85 & 0.84 & 0.85 & 0.86 & 0.86 & 0.43 & 0.84 & 0.84 & 0.99\\
\hline
Chinatown & 0.95 & 0.95 & 0.95 & 0.92 & 0.95 & 1.0 & 0.89 & 0.92 & 1.0 & 1.0\\
\hline
CBF & 0.97 & 0.67 & 1.0 & 0.97 & 0.7 & 0.91 & 0.95 & 0.97 & 1.0 & 1.0\\
\hline
InsectEPGRegularTrain & 0.87 & 0.75 & 0.98 & 0.91 & 0.65 & 0.99 & 0.91 & 0.98 & 0.98 & 1.0\\
\hline
TwoLeadECG & 0.81 & 0.98 & 0.93 & 0.84 & 0.98 & 0.88 & 0.73 & 0.88 & 0.95 & 1.0\\
\hline
ItalyPowerDemand & 0.94 & 0.8 & 0.94 & 0.94 & 0.83 & 0.92 & 0.67 & 0.92 & 0.94 & 1.0\\
\hline
InsectEPGSmallTrain & 0.87 & 0.85 & 0.87 & 0.87 & 0.64 & 0.87 & 0.9 & 0.87 & 0.87 & 1.0\\
\hline
ToeSegmentation2 & 0.85 & 0.8 & 0.97 & 0.83 & 0.87 & 0.86 & 0.92 & 0.99 & 0.93 & 1.0\\
\hline
BeetleFly & 0.89 & 0.86 & 0.92 & 0.89 & 0.95 & 0.71 & 0.86 & 0.86 & 0.86 & 1.0\\
\hline
Herring & 0.66 & 0.65 & 0.68 & 0.61 & 0.7 & 0.58 & 0.68 & 0.65 & 0.65 & 0.99\\
\hline
ECGFiveDays & 0.79 & 0.78 & 0.86 & 0.79 & 0.78 & 0.9 & 0.85 & 0.89 & 0.87 & 0.93\\
\hline
DistalPhalanxOutlineCorrect & 0.85 & 0.84 & 0.86 & 0.85 & 0.83 & 0.85 & 0.51 & 0.86 & 0.87 & 1.0\\
\hline
Fish & 0.85 & 0.9 & 0.93 & 0.82 & 0.91 & 0.86 & 0.26 & 0.92 & 0.91 & 1.0\\
\hline
ToeSegmentation1 & 0.67 & 0.53 & 0.73 & 0.64 & 0.47 & 0.62 & 0.86 & 0.84 & 0.75 & 0.99\\
\hline
Meat & 0.97 & 0.89 & 0.97 & 0.96 & 0.89 & 0.99 & 0.5 & 0.97 & 0.98 & 1.0\\
\hline
Trace & 0.91 & 0.94 & 0.94 & 0.89 & 0.95 & 0.88 & 0.84 & 0.91 & 0.93 & 1.0\\
\hline
Symbols & 0.7 & 0.83 & 0.84 & 0.7 & 0.68 & 0.76 & 0.68 & 0.83 & 0.83 & 1.0\\
\hline
Ham & 0.79 & 0.84 & 0.89 & 0.8 & 0.82 & 0.9 & 0.68 & 0.9 & 0.91 & 1.0\\
\hline
GunPointOldVersusYoung & 0.91 & 0.86 & 1.0 & 0.89 & 0.86 & 0.95 & 1.0 & 1.0 & 0.96 & 1.0\\
\hline
ShapeletSim & 0.67 & 0.62 & 0.95 & 0.79 & 0.62 & 0.52 & 1.0 & 0.92 & 0.95 & 1.0\\
\hline
HouseTwenty & 0.79 & 0.81 & 0.87 & 0.92 & 0.87 & 0.77 & 0.96 & 0.87 & 0.96 & 1.0\\
\hline
OSULeaf & 0.81 & 0.91 & 0.89 & 0.83 & 0.92 & 0.77 & 0.81 & 0.88 & 0.82 & 1.0\\
\hline
DistalPhalanxTW & 0.83 & 0.83 & 0.84 & 0.83 & 0.83 & 0.86 & 0.75 & 0.86 & 0.84 & 1.0\\
\hline
Wine & 0.91 & 0.92 & 0.95 & 0.91 & 0.92 & 0.94 & 0.69 & 0.95 & 0.95 & 1.0\\
\hline
UMD & 0.81 & 0.84 & 0.82 & 0.79 & 0.71 & 0.82 & 0.65 & 0.8 & 0.85 & 1.0\\
\hline
Fungi & 0.0 & 0.0 & 0.0 & 0.0 & 0.0 & 0.0 & 0.0 & 0.0 & 0.0 & 1.0\\
\hline
ProximalPhalanxOutlineCorrect & 0.88 & 0.9 & 0.88 & 0.88 & 0.9 & 0.89 & 0.81 & 0.88 & 0.89 & 0.99\\
\hline
Mallat & 0.97 & 0.86 & 0.96 & 0.99 & 0.94 & 0.99 & 0.68 & 0.95 & 0.97 & 1.0\\
\hline
SwedishLeaf & 0.82 & 0.9 & 0.94 & 0.82 & 0.89 & 0.82 & 0.36 & 0.9 & 0.89 & 1.0\\
\hline
GunPointAgeSpan & 0.89 & 0.83 & 0.98 & 0.88 & 0.84 & 0.97 & 0.96 & 0.98 & 0.97 & 1.0\\
\hline
ECG5000 & 0.9 & 0.91 & 0.96 & 0.91 & 0.91 & 0.96 & 0.73 & 0.96 & 0.96 & 1.0\\
\hline
Plane & 0.96 & 0.97 & 0.96 & 0.95 & 0.96 & 0.94 & 0.84 & 0.95 & 0.96 & 1.0\\
\hline
MoteStrain & 0.84 & 0.74 & 0.86 & 0.84 & 0.74 & 0.85 & 0.92 & 0.86 & 0.86 & 1.0\\
\hline
Beef & 0.39 & 0.51 & 0.49 & 0.41 & 0.5 & 0.51 & 0.34 & 0.49 & 0.5 & 0.97\\
\hline
SonyAIBORobotSurface1 & 0.88 & 0.93 & 0.94 & 0.88 & 0.97 & 0.91 & 0.71 & 0.91 & 0.89 & 1.0\\
\hline
DistalPhalanxOutlineAgeGroup & 0.86 & 0.85 & 0.89 & 0.85 & 0.85 & 0.87 & 0.65 & 0.88 & 0.89 & 0.99\\
\hline
ArrowHead & 0.73 & 0.91 & 0.89 & 0.75 & 0.91 & 0.91 & 0.62 & 0.89 & 0.89 & 0.99\\
\hline
DiatomSizeReduction & 0.64 & 0.93 & 1.0 & 0.64 & 0.9 & 0.97 & 0.48 & 1.0 & 1.0 & 1.0\\
\hline
Coffee & 0.94 & 0.98 & 1.0 & 0.94 & 0.98 & 1.0 & 0.67 & 1.0 & 1.0 & 1.0\\
\hline
ProximalPhalanxOutlineAgeGroup & 0.86 & 0.88 & 0.87 & 0.86 & 0.87 & 0.88 & 0.52 & 0.86 & 0.86 & 0.99\\
\hline
BME & 0.82 & 0.89 & 0.87 & 0.87 & 0.85 & 0.93 & 0.68 & 0.89 & 0.82 & 1.0\\
\hline
SmoothSubspace & 0.91 & 0.88 & 0.99 & 0.92 & 0.87 & 0.95 & 0.5 & 0.99 & 1.0 & 1.0\\[2pt]
\hline
\end{tabular}\label{tab:64results}}
\end{center}
\end{table}

In all but seven datasets, the DGAP distances obtained the highest F1 scores, so that DGAP strictly outperformed the other distances approximately $89\%$ of the time. In the remaining seven datasets, the DGAP F1 scores were tied with others which obtained scores of $1.0$: therefore, DGAP was never outperformed. For reference, the names of the UCR datasets for which ties occurred for the highest F1 score are as follows:

\begin{itemize}
    \item GunPointMaleVersusFemale
    \item Chinatown
    \item CBF
    \item GunPointOldVersusYoung
    \item ShapeletSim
    \item DiatomSizeReduction
    \item Coffee
\end{itemize}

The high F1 scores for DGAP demonstrate a possible strong tie between points which are misclassified by the proximity forest algorithm and points that may be considered outliers based on the forest proximities. The connection between misclassified points and outliers appears to be stronger than for $1$-NN classifiers using several commonly-used time series distance measures.

\begin{figure*}[htp]
  \centering
  \subfigure[MDS embedding of the \textit{GunPoint} dataset using pairwise dissimilarity defined by dynamic time warping distances.]{\includegraphics[scale=0.35]{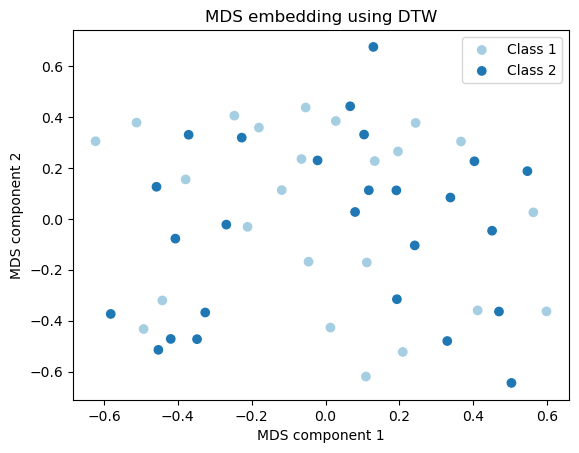}}\label{GunPointFigA}\qquad
  \subfigure[MDS embedding of the \textit{GunPoint} dataset using pairwise dissimilarities defined by PF-GAP. The point with the highest within-class outlier score is indicated in red and belongs to Class 2.]{\includegraphics[scale=0.35]{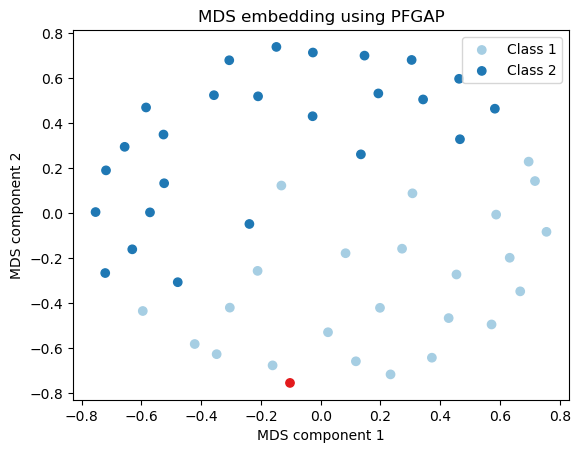}}\label{GunPointFigB}
  \caption{MDS embeddings of the \textit{GunPoint} dataset.}
  \label{GunPointFig}
\end{figure*}

This connection can also be visualized, since the distances defined in equation \ref{dissim2} can be coupled with various manifold learning methods to produce a vector space embedding of the time series data. Additionally, since the distances have been computed in a supervised manner, the resulting embedding will tend to provide better class separation when compared with embeddings based on unsupervised distances. In Figure \ref{GunPointFig}, 2-dimensional embeddings obtained using MDS are shown, offering a visual comparison of class separation obtained using distances defined by $p_{GAP}$ as opposed to distances defined in an unsupervised manner. We note, however, that RF proximities have also been embedded using other methods \cite{Vaiciukynas2017parkinsons,RFPHATE,rhodes2023gaining,rhodes2024rfma}.

As previously shown with tabular data, distances defined by $p_{GAP}$ visually reflect the classification strength of the underlying model, and within-class outliers tend to also be visual outliers \cite{RFGAP}. Figure \ref{GunPointFig} appears to suggest that the same desirable properties tend to apply for PF-GAP.

\section{Discussion of Limitations}
\label{sec:Discussion}

The dissimilarity defined by $p_{GAP}$ outperformed or tied in with the dissimilarities defined by commonly used time series distance measures for the 64 UCR datasets used. 
The proximity forest classifier has been shown to generally outperform $1$-NN classifiers \cite{PF}, which is a conceivable contribution to the higher F1 scores for classification/inlier association. However, the F1 scores allow for weak classifiers, provided that the associated distance measure, coupled with LOF, tends to label misclassified points as outliers.

One possible limitation in our method is the formula we have used to define dissimilarities from $p_{GAP}$, given in Equation \ref{dissim2}. We have advocated for the use of the exponent value of $2$ on the basis that the proximities are typically small. However, it is possible that using a different formula to define dissimilarity from $p_{GAP}$ proximities will have a positive or negative impact on subsequent tasks, including obtaining time series embeddings and outlier detection.

There may also be limitations imposed by the proximity forest classifier itself. The number of proximity trees chosen is a hyperparameter and is likely to have a large impact on the quality of the $p_{GAP}$ proximities. In particular, random forests are commonly trained with several hundred trees, whereas a tree count of less than $100$ trees is not uncommon for a proximity forest. When fewer trees are chosen, the likelihood of points being in-bag for every tree increases, affecting the proximities defined by $p_{GAP}$.

\section{Conclusion}
\label{sec:Conclusion}

We have introduced forest proximities for the PF model, specifically $p_{GAP}$ proximities previously defined for random forests \cite{RFGAP}, which we term PF-GAP. Using the forest proximities, we have demonstrated how to construct time series vector embeddings, which may produce better inter-class separation than unsupervised time series distance-induced embeddings. We have also introduced intraclass outlier scores for time series based on forest proximities. 
We have used the forest proximities along with LOF to detect outliers, showing a strong connection between points which are misclassified by a proximity forest classifier and points which are considered outliers using the forest proximities. Future work includes further study of the use of PF-GAP in time series outlier detection--that is, in the detection of within-class anomalous time series.


Future work also includes the exploration of additional applications of forest proximities. With random forests, forest proximities have been used directly in classification without the need to predict using the forest structure \cite{RFGAP}. This and other potential applications of forest proximities for proximity forests we leave as a future endeavor.
Additionally, a ``proximity forest 2.0'' algorithm has recently been introduced \cite{PF2}, promising improved computational speed and classification accuracy over the original proximity forest algorithm. Thus, it is natural to consider the future work of calculating forest proximities for proximity forest 2.0, as well as for other forest-based time series classifiers. 

\bibliographystyle{IEEEtran}
\bibliography{mybib}

\end{document}